# A Deep Reinforcement Learning Approach for Online Parcel Assignment

Hao Zeng
Cainiao Network
Hangzhou, China
zenghao.zeng@cainiao.com

Qiong Wu
Cainiao Network
Hangzhou, China
melody.wq@cainiao.com

Kunpeng Han
Cainiao Network
Hangzhou, China
kunpeng.hkp@cainiao.com

Junying He
Cainiao Network
Hangzhou, China
junying.hjy@cainiao.com

Haoyuan Hu
Cainiao Network
Hangzhou, China
haoyuan.huhy@cainiao.com

## ABSTRACT

In this paper, we investigate the online parcel assignment (OPA) problem, in which each stochastically generated parcel needs to be assigned to a candidate route for delivery to minimize the total cost subject to certain business constraints. The OPA problem is challenging due to its stochastic nature: each parcel's candidate routes, which depends on the parcel's origin, destination, weight, etc., are unknown until its order is placed, and the total parcel volume is uncertain in advance. To tackle this challenge, we propose the PPO-OPA algorithm based on deep reinforcement learning that shows competitive performance. More specifically, we introduce a novel Markov Decision Process (MDP) framework to model the OPA problem, and develop a policy gradient algorithm that adopts attention networks for policy evaluation. By designing a dedicated reward function, our proposed algorithm can achieve a lower total cost with smaller violation of constraints, comparing to the traditional method which assigns parcels to candidate routes proportionally. In addition, the performances of our proposed algorithm and the Primal-Dual algorithm are comparable, while the later assumes a known total parcel volume in advance, which is unrealistic in practice.

## 1 INTRODUCTION

The online parcel assignment (OPA) problem naturally arises from today's e-commerce environment, where the logistics company needs to assign each parcel to a candidate route for delivery after customers make online purchases. As shown in Figure 1, a candidate route consists of multiple logistics service providers and physical nodes such as hubs. When a parcel is assigned to a candidate route, it consumes resource of all providers and hubs within the route, and raises a delivery cost to be paid by the logistics company. The set of available candidate routes and their corresponding delivery costs are determined by the parcel's attributes such as origin, destination, weight, etc., which remain unknown until the parcel order is made. As online shopping prevails and daily parcel volume grows tremendously, it becomes crucial for the logistics company to make parcel assignments wisely because it significantly influences the total delivery cost. Other than delivery cost, business constraints due to resource capacities or established contracts, need to be considered in this problem. A business constraint can be interpreted as the lower and upper bounds of the number of parcels that can be assigned to a provider or hub. The OPA problem is to assign each stochastically generated parcel to a candidate route with the objective as minimizing the total delivery cost subject to given business constraints.

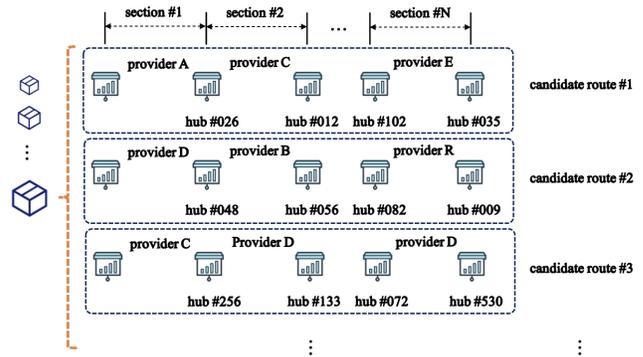

Figure 1: Incoming parcels and their candidate logistics routes in online parcel assignment.

The OPA problem is closely related to several problems that have been studied in the literature. By assuming all incoming parcels' attributes and candidate routes are known, the offline version of the parcel assignment problem can be formulated as a deterministic integer programming problem. If we assume the total parcel volume to be assigned is given, which is not true in real practice, our problem would be similar to the online allocation problem [6, 31]. In the setting of the online allocation problem, the total number of requests is assumed to be given, but the arrival sequence of requests are unknown. Such problem appears in many practices such as adwords matching [11, 18], online routing[7] and online combinatorial auction[9]. For the online allocation problem, there exists a competitive ratio of $(1 - 1/e)$ with adversarial arrivals and $(1 - \varepsilon)$ with stochastic arrivals [8]. When dropping out of the assumption of known total number of requests, the OPA online problem can be viewed as a special kind of the online resource allocation problem under horizon uncertainty[5], where the horizon tends to be large ($> 10^5$ parcels to be assigned per day) but remains unknown until the end of the decision-making process. Besides, the attribution of future parcels (including origin, destination, and weight etc) is unknown and irrelevant to assignment policy, which violates the Markovian property. Therefore, in this paper, we focus on solving the OPA problem utilizing a DRL method. By using a modified MDP to formulate the OPA problem, we propose a proximal policy optimization algorithm, in which assignment decisions are made based on current observation and past information, to optimize the objective while keeping the violation of constraints as small as possible.

Several methods have been attempted for solving the OPA problem. For example, the greedy method, which always assigns the parcel to the route with the lowest cost, can guarantee to minimize the total cost. However, since this method does not take any constraint into account, the possibly severe violation of constraints makes it inappropriate for real practice. Other typical algorithms based on online primal-dual framework [8] has been used to solve a variety of online optimization problem, such as online adwords problem [11], online task assignment in crowd-sourcing markets [15]. Nevertheless, the online primal-dual algorithm requires the total number of parcels given, which is impractical for actual scenarios. Recently, deep reinforcement learning (DRL) approaches have received great attention for their capacity to solve complex decision-making problems efficiently. In this paper, we propose the PPO-OPA algorithm, showing that the DRL approach can be powerful for solving the OPA problem.

The main contributions of this paper are summarized as follows:

- We propose an Online Assignment Markov Decision Process framework for modeling the decision-making process in online parcel assignment situations, which can also be applied to a variety of online assignment problems.
- Based on the PPO framework, we propose a DRL method that uses attention networks to learn the feature combination of incoming parcel's information and constraints' status for improving the assignment policy.
- In the experiments, we test our proposed PPO-OPA algorithm using real datasets from Cainiao Network. The results show that our approach outperforms the traditional assignment method used in the logistics industry. In addition, we show that the performances of our proposed algorithm and the Primal-Dual algorithm are comparable.

## 2 RELATED WORK

Our algorithm and analysis build on the Markov Decision Process (MDP) framework, which provides a widely applicable mathematical formalism in sequential decision-making problems. In the MDP framework, the agent observes a state $s_t$ from the environment at each time step $t$, and then makes an action according to its policy $\pi(s_t)$. After an action is taken, the state transits to the next state $s_{t+1}$ and a reward $r_t$ is sent back from the environment to the agent. The goal of MDP is maximizing accumulated discounted reward $R = \sum_{t=1}^{T} \gamma^{t-1} r_t$ by learning an optimal policy, where $\gamma \in (0, 1]$. A general algorithm for solving MDP is reinforcement learning [25] in which well-trained agent takes actions in an environment in order to maximize the cumulative reward. Recently, deep reinforcement learning (DRL) methods employ neural networks for function approximation [19] to handle high-dimensional state and action space. The most successful achievements include AlphaGo [24] and AlphaZero [23], which convincingly defeated world champion programs in chess, Go, and Shogi, without any domain knowledge other than underlying rules as input during training.

The commonly useful DRL methods mainly include value learning and policy gradient. Value learning are aimed at explicit learning of value functions from which the optimal policy can be obtained. A commonly used branch of value learning includes Deep Q-Network (DQN) [19] and its variants (e.g., Rainbow [14]) are mostly suitable for discrete action space and are successful in mastering a range of Atari 2600 games. The policy gradient methods, meanwhile, attempt to learn optimal policies directly. Policy gradient methods with the assistance of baselines (e.g., value functions) are also referred to as Actor-Critic methods, which are suitable for both discrete and continuous action space. Representative Actor-Critic methods are (DDPG) [17], TRPO [21] and PPO [22] etc. TRPO develops a series of approximations and the original objective of policy gradient is converted to minimization of a surrogate loss function with the constraint of KL divergence between old and new policy, which use the trust-region method to guarantee policy improvement with non-trivial step sizes. PPO [22] is a substitute of TRPO, which is more applicable to large-scale decision problems. The algorithms mentioned above rely on basic assumptions that the problem can be model to a MDP, which makes some online problems not be applied. To bridge between reinforcement learning and online learning, Even-Dar et al. [13] proposed online MDP that relaxed the Markovian assumption of the MDP by setting the reward function be time dependent. Similarly, our work extends Markov Decision Process to match online parcel assignment problem by introducing uncertain observation at each step. This method can efficiently employ the exploration-exploitation benefits of RL algorithms to acquire an effective policy.

There have been an increasing number of studies on employing DRL methods for industrial decision-making problems. Zhang and Diettterich [32] utilized temporal difference learning $TD(\lambda)$ to learn a heuristic evaluation function over states to learn domain-specific heuristics for job-shop scheduling. Tesauro et al.[26, 27] showed the feasibility of online RL to learn resource valuation estimates which can be used to make high-quality server allocation decisions in multi-application prototype data center scenarios. Recently, Ye Li and Juang [30] develop a novel decentralized resource allocation mechanism for vehicle-to-vehicle (V2V) communications based on DRL. In order to reach the objective of minimizing power consumption and meeting the demands of wireless users over a long operational period, Xu et al. [29] present a novel DRL-based framework for power-efficient resource allocation in cloud RANs. Du et al. [12] learn a policy that maximizes the profit of the cloud provider through trial and error. They integrate long short-term memory (LSTM) neural networks into improved DDPG to deal with online user arrivals, which addresses both resource allocation and pricing problems. In summary, most of these studies assume that the environment is Markovian. Nevertheless, the OPA problem violates Markovian assumption such that DRL algorithm is hardly used for solving online parcel assignment problems directly.

Another classic algorithm building on primal-dual framework, has already been applied in a variety of online optimization problems [4, 8]. For example, online adwords problem with stochastic assumption, where keywords arrive online, and advertisers must be assigned to keywords such that the revenue is maximum without exceeding any advertiser's budget. The online primal-dual framework can achieve near-optimal performance under the case where the total budget is sufficiently large [3]. In addition, Ho and Vaughan [15] introduced the online task assignment problem in crowdsourcing markets, in which workers arrive one at a time, and must be assigned to a task. By designing Dual Task Assigner based on the primal-dual framework, this work proved DTA outperforms other algorithms empirically. However, the DTA algorithm also requires the total number of workers given in advance, which is difficult to obtain in practicality. Recently, Balseiro et al. [5] combined dual descent with a carefully-chosen target consumption sequence to solve online resource allocation under horizon uncertainty, which do not require knowing the number of requests, and proved that it achieves a bounded competitive ratio when the horizon uncertainty grows large.

## 3 PROBLEM FORMULATION

We now formulate the online parcel assignment problem. During a period of time, we define the total parcel volume as $m$. Online parcel assignment problem requires assigning each incoming parcel to one of its candidate routes to minimize the total cost subject to the constraints of service providers. We use $\mathcal{J}(i)$ to define the set of all candidate routes of the parcel $i$. Let $\mathcal{K}$ be the set of all constraints. $C(i, k)$ denote the set of routes corresponding parcel $i$ and constraint $k$. Besides, we define decision variables $x_{i,j}$ which is 1 if parcel $i$ is assigned to routes $j$ and corresponding cost $c_{i,j}$. Therefore, the offline parcel assignment problem can be formulated to the following linear programming

$$\min_{x} \sum_{i=0}^{m} \sum_{j \in \mathcal{J}(i)} c_{ij} x_{ij}$$
$$\text{s.t.} \sum_{j \in \mathcal{J}(i)} x_{ij} = 1, \quad \forall i \quad (1)$$
$$L_k \leq \sum_{j \in C(i,k)} \sum_{i=0}^{m} x_{ij} \leq U_k, \quad \forall k \in \mathcal{K}$$
$$x_{ij} \in \{0, 1\}, \quad \forall j \in \mathcal{J}(i), \forall i$$

In this paper, we mainly consider two types of constraints as follows:

- **Capacity Constraints:** the upper bound of parcel volume for providers' hub.
- **Proportion Constraints:** For each pair of origin and destination, the percentage of parcels served by some providers needs to be within given ranges.

For Capacity Constraints, $L_k = 0$ and $U_k$ is the upper bound capacity of hub $k$. For Proportion Constraints, $L_k = p_k^L \cdot n_k, U_k = p_k^U \cdot n_k$, where $p_k^L$ and $p_k^U$ is given by providers and $n_k$ denote the number of parcels corresponding to constraint $k$. It should be noted that $n_k$ is known under the offline setting but unknown in the online context due to unidentified upcoming parcels.

In an online setting, the parcel assignment problem has an unknown total parcel volume $m$ and unpredictable future parcel information, that is non-Markovian. The agent needs to make decisions based on the incoming parcel attributions and constraint information, however, only constraint states satisfy Markov property. Therefore, it is necessary to extend MDP for reinforcement learning algorithms that can be used for online assignment problem. We offer the following definition of Online Assignment MDP by introducing observations $O$,

DEFINITION 1 (ONLINE ASSIGNMENT MDP). *An online assignment MDP is a 5-tuple $(O, S, \mathcal{A}, \mathcal{P}, \mathcal{R})$, where*

- $O$ *is a set of observations from an unknown distribution,*
- $S$ *is a set of states,*
- $\mathcal{A}$ *is a set of actions,*
- $\mathcal{P}$ *is a state transition probability*
  $$\mathcal{P}_{o,s,s'}^{t} = \Pr(S_{t+1} = s' | O_t = o, S_t = s, A_t = a),$$
- $\mathcal{R}$ *is a reward function,*
  $$\mathcal{R}_{o,s}^{a} = \mathbf{E}(R_{t+1} | O_t = o, S_t = s, A_t = a).$$

Generally, Definition 1 can also be applied to model a variety of online allocation problems under unknown upcoming requests. For example, $O_t$ can denote the online arrival keyword at time $t$ and online arrival worker at time $t$ for online adwords problems and online task assignment problems respectively. One has to notice that Online Assignment MDP is different from Partially Observable MDP whose observation depends on the new state or action, so it is necessary to propose new MDP definition for solving the OPA problem. We now formulate the specific components for online parcel assignment problem at time $t$:

- **Observation**: incoming parcel information $o_t$,
- **State**: Current constraint status $s_t$. For capacity constraints, $s_t = \{h_i(t)/h_i, i \in \mathcal{H}\}$, where $\mathcal{H}$ is the set of all hubs and $h_i$ is the upper bound of capacity for hub $i$. For proportion state, $s_t = \{p_j(t), j \in \mathcal{R}\}$, where $\mathcal{R}$ is the set of all routes and $p_j(t)$ is the current ratio for the providers in route $j$.
- **Action**: action sample from a discrete distribution corresponding each candidate routes of parcel $o_t$.
- **Reward**: The Design of the reward is the most challenging part of the problem. At each time step, the immediate reward should integrate constraint state and parcel information. Since the objective is to minimize the total cost, the first part of the reward is the cost of the assigned route $c_{a_t}$, which depends on action $a_t$. For capacity constraints, the smaller the remaining capacity, the greater the penalty. For proportion constraints, we will encourage making the proportion close to the lower bound if the current proportion is less than it. Otherwise, we will give punishment if the current proportion is greater than the upper bound. Hence, the reward is designed as follows:

$$r_t = -c_{a_t} + \lambda f_{a_t}(t), \quad (2)$$

where $\lambda$ is a hyperparameter to leverage the importance of constraint state function $f_{a_t}(t)$ and cost $c_{a_t}$. If constraint $i$ is the capacity constraint, then

$$f_i(t) = e^{-h_i(t)/h_i}; \quad (3)$$

if constraint $i$ is proportion constraint, then

$$f_i(t) = I(p_i(t) < L_i)(L_i - p_i(t)) \\ - I(p_i(t) > U_i)(p_i(t) - U_i)), \quad (4)$$

where $I(\cdot)$ is the indicator function.

Another class method for reward design is to use negative cost as reward directly and use constrained MDP (CMDP) method [1, 10] to solve. In the experiment, we combined Lagrangian relaxation with proposed DRL algorithm to control constraint violation, however, it can not achieve a better performance compared to methods that add a penalty to reward for online parcel assignment problems.

## 4 PPO-OPA ALGORITHM

In this section, we propose a DRL method based on Proximal policy optimization (PPO) [22] for solving the OPA. PPO is a commonly used RL algorithm with excellent performance for solving a variety of MDP problems. As an Actor-Critic algorithm, the policy function and state value function (often represented by actor network and critic network) need to be estimated during training. PPO adopts the advantage function to assist update gradient and reduce variance. One common useful advantage function $A_t$ viewed as its temporal-difference (TD) error estimation utilized in value functions estimation is:

$$A_t = r_t + \gamma V(s_{t+1}) - V(s_t).$$

However, for online parcel assignment problem, value function also depends on incoming parcel information, that is $V(o_t, s_t)$, while $o_t$ is non-Markovian and difficult to predict, so does $V(o_t, s_t)$. Therefore, replacing to estimate future accumulated reward (by critic network), we use reward network $R_\phi(s_t, o_t)$ as the estimator of reward based on the current state and incoming parcel. Then, we define the advantage $A_t$ can be defined as the following

$$A_t = r_t - R_\phi(s_t, o_t), \quad (5)$$

where $R_\phi(\cdot, \cdot)$ is the reward function which is represented by the reward network. The loss function for this reward network update is

$$E_\phi \left[ (R_\phi(s_t, o_t) - r_t)^2 \right]. \quad (6)$$

Parcel information includes the parcel attribution features and candidate features. One parcel corresponds to multiple candidate routes and different parcels may have different numbers of candidate routes. Therefore, we propose actor network to catch parcel and route features as shown in Figure 2. Parcel features and candidate-route features are respectively inputted to different embedding layers with Multiple Layer Perceptron (MLP) afterward. For the candidate-route feature, we use identical parameters for each route. We define $N_R$ as the maximum number of possible candidate routes per parcel. If the number of candidate routes is less than $N_R$ we will construct fictitious routes with default costs and default constraint states, of which the values are set to be all 0. Then, a mask matrix is used to convert the output of fictitious routes to 0.

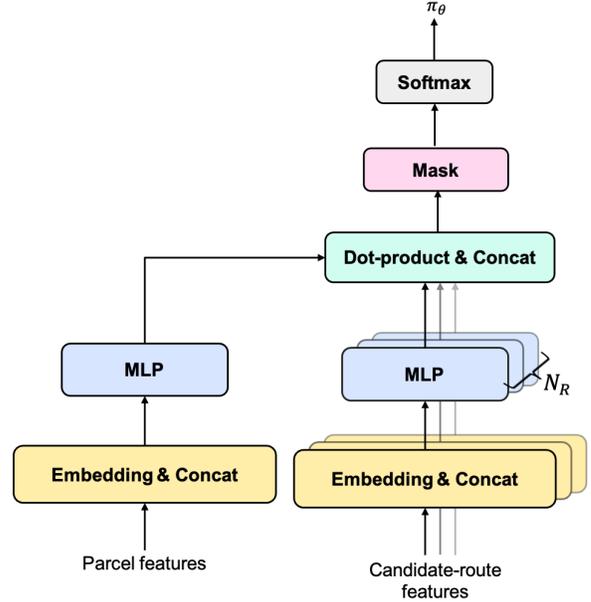

Figure 2: The actor network. Parameter sharing is applied to route vectors in candidate-route features and probabilities of assigning each route are output from the softmax layer.

The reward network in Figure 3 is similar to the actor network. Parameter sharing is also utilized to accommodate candidate-route features. Besides, attention mechanism [28] is employed to calculate the state value for each incoming parcel. An attention function can be described as mapping a query and a set of key-value pairs to an output. The output is computed as a weighted sum of the values, where the weight assigned to each value is computed by a compatibility function of the query with the corresponding key. We treat parcel features as a query, and candidate-route features are used to generate key and value, which is

$$q = W^q o, k_i = W^k h_i, \text{ and } v_i = W^v h_i, \quad (7)$$

where $i \in [1, 2, ..., N_R]$. Then, the weighted sum $v$ is computed by

$$v = \sum_{i=1}^{n} q^T k_i v_i. \quad (8)$$

Finally, reward value can be obtained by

$$R_\phi(s_t, o_t) = \text{MLP}(v). \quad (9)$$

Our network design is very simple, which guarantees to infer rapidly in industrial applications. The experimental results show the effectiveness of the network structure.

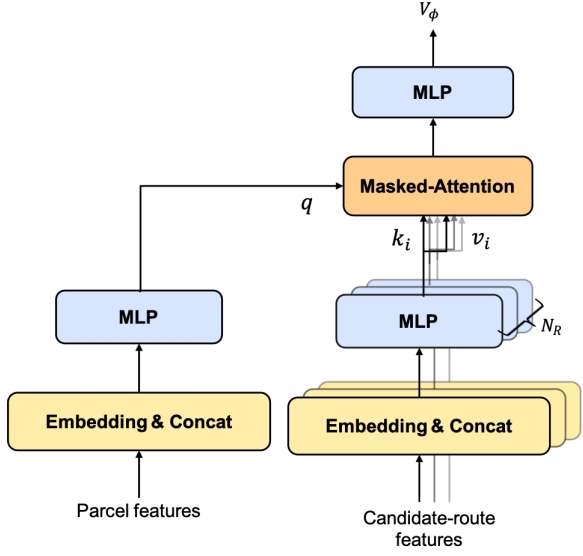

Figure 3: The reward network. Parameter sharing is applied to route vectors in candidate-route features. Masked-Attention layers are used for calculating state value function given a certain parcel and current constraint state.

Therefore, based on PPO algorithm, the improved clipped optimization objective for policy updating is

$$L^{\text{CLIP}}(\theta; \hat{\pi}) = E_{\tau \sim \hat{\pi}}\left[\sum_{t=0}^{T} \min(p_t(\theta; \hat{\pi}), \text{CLIP}(p_t(\theta; \hat{\pi}), 1-\varepsilon, 1+\varepsilon))A_t\right], \quad (10)$$

where $p_t(\theta; \hat{\pi}) = \frac{\pi_\theta(a_t|s_t,o_t)}{\hat{\pi}(a_t|s_t,o_t)}$. Our DRL algorithm is proposed in Algorithm 1. The trajectories are collected in parallel through policy $\pi_{\theta_k}$ (line 2). Then the network parameters $\theta$ and $\phi$ are updated by using Adam [16].

## 5 PERFORMANCE EVALUATION

We implement and evaluate the PPO-OPA algorithm on a workstation computer (ubuntu 16.04), which is has an Intel Xeon Platinum 8163 @ 2.50 GHz, 32 GB memory and an Nvidia Tesla V100 GPU with 16 GB memory. We use PyTorch [20] for implementation. For the neural network setting, we set the embedding dimensions as 64 in both the actor and reward networks. For the actor network, the MLP part for parcel features has a single layer of 128 neurons, while that for candidate route features has two layers: one has 256 neurons and the other has 128 neurons. All layers are with ReLu activation. For the reward network, the MLP parts have the same settings as those in the actor network. After the masked-attention layer, the last MLP part has a layer with 64 neurons with Sigmoid activation followed by a linear layer with 1 neuron. The learning rates for the actor and reward networks are set to be $10^{-3}$ and $10^{-3}$, respectively. The importance hyperparameters $\lambda$'s are set to be 10 and 300 for the capacity and proportion constraints, respectively.

---

**Algorithm 1** Proximal policy optimization for online parcel assignment (PPO-OPA)

**Input**: initial policy parameters $\theta_0$ and initial value function parameters $\phi_0$.

1: **for** $k = 0, 1, 2, \ldots$ **do**
2:     Collect set of trajectories $D_k = \{\tau_i\}$ by running policy $\pi_{\theta_k}$ in the environment.
3:     Compute rewards $r_t$ for each trajectory.
4:     Compute advantage estimates, $A_t = r_t - R_{\phi_k}(s_t, o_t)$.
5:     Update policy by maximizing the PPO objective:

$$\pi_{k+1} = \arg\max_{\theta} \frac{1}{|D_k|T} \sum_{\tau \in D_k} L^{\text{CLIP}}(\theta, \pi_{\theta_k}),$$

    typically via stochastic gradient descent with Adam.
6:     Fit value function by regression on mean-squared error:

$$\phi_{k+1} = \arg\min_{\phi} \frac{1}{|D_k|T} \sum_{\tau \in D_k} \sum_{t=0}^{T} \left(R_\phi(s_t, o_t) - r_t\right)^2,$$

    typically via stochastic gradient descent with Adam.
7: **end for**

---

**Datasets**: we use two datasets, denoted as dataset #1 and dataset #2, both of which are real data provided by Cainiao Network. Each dataset contains two parts of data, namely parcel data and constraints configuration data.

- Parcel data: This data contains the records of historical parcels created within a country and a time period, sorted by their creation times. Each record shows one parcel's information, including its candidate routes and corresponding costs.
- Constraints configuration data. This data contains the configuration of business constraints, such as capacity constraints and proportion constraints, that should be considered while making the assignments for those in the parcel data.

Dataset #1 contains 625 hub capacity constraints and the daily parcel volume varies from 567429 to 806824. On the other hand, dataset #2 contains only 51 proportion constraints and the daily parcel volume is smaller, ranging from 293208 to 326332.

In the training procedure, we select the parcel data of datasets #1 and #2 created within a particular day $T$. That is, 684793 records from dataset #1 and 308329 from dataset #2 are selected. The agent uses this data for trajectory collection and trains neural networks about 20 episodes for attaining convergence. In each episodes, we first collect 50 trajectories in parallel and put all MDP tuples in the trajectories into a buffer. Then, we shuffle the buffer and update the parameters of the actor and reward networks using Adam. The mini-batch size for gradient descent is 2048. In the validation procedure, we use the parcel data of datasets #1 and #2 created within the next three days (i.e., $T+1$, $T+2$, $T+3$).

We compare the results from PPO-OPA against those from three other online algorithms and the integer programming (IP) method, descriptions of which are as follows:

(1) **IP**: the OPA problem can be formulated as an IP problem (1), if all parcels are known in advance. It is straightforward that the solution to (1) is optimal for the OPA problem. Therefore, we can use the IP gap, the difference between the optimal objective value and the objective value from certain algorithm, as a measurement of performance. To solve (1), we use SCIP [2], a commonly used solver for IP problems.

| Dataset #1 | Algorithm | Average Cost | IP Gap | Violation Rate |
|---|---|---|---|---|
| | PPO-OPA | 100.73 | 0.0688% | 2.53% |
| | PPO-PD | 102.05 | 1.3834% | 4.20% |
| $T+1$ | Proportion | 101.05 | 0.3874% | 2.50% |
| | PDO | 100.67 | 0.0671% | 2.50% |
| | IP(offline) | 100.66 | | |
| | PPO-OPA | 99.782 | 0.0662% | 6.32% |
| | PPO-PD | 101.24 | 1.5242% | 8.95% |
| $T+2$ | Proportion | 100.19 | 0.4723% | 6.28% |
| | PDO | 99.719 | 0.0030% | 6.26% |
| | IP(offline) | 99.716 | | |
| | PPO-OPA | 98.479 | 0.0872% | 5.44% |
| | PPO-PD | 100.47 | 2.1148% | 9.33% |
| $T+3$ | Proportion | 98.927 | 0.5457% | 5.37% |
| | PDO | 98.390 | -0.0027% | 5.39% |
| | IP(offline) | 98.393 | | |

Table 1: The evaluation results in dataset #1 with capacity constraints. The agent is trained by using the parcel data from day $T$. The parcel volumes for $T+1, T+2, T+3$ are 567429, 756579 and 806824 respectively.

| Dataset #2 | Algorithm | Average Cost | IP Gap | Violation Rate |
|---|---|---|---|---|
| | PPO-OPA | 81.193 | -0.1276% | 2.57% |
| | PPO-PD | 81.130 | -0.2052% | 5.54% |
| $T+1$ | Proportion | 81.459 | 0.1993% | 3.39% |
| | PDO | 81.139 | -0.1946% | 5.37% |
| | IP(offline) | 81.297 | | |
| | PPO-OPA | 78.565 | 0.0495% | 3.31% |
| | PPO-PD | 78.472 | -0.0685% | 8.00% |
| $T+2$ | Proportion | 78.723 | 0.2509% | 4.87% |
| | PDO | 78.493 | -0.0419% | 2.42% |
| | IP(offline) | 78,526 | | |
| | PPO-OPA | 84.753 | 0.1213% | 2.25% |
| | PPO-PD | 84.659 | 0.0105% | 6.95% |
| $T+3$ | Proportion | 84.930 | 0.3308% | 3.31% |
| | PDO | 84.683 | 0.0392% | 2.06% |
| | IP(offline) | 84.650 | | |

Table 2: The evaluation results for dataset #2 with proportion constraints. The agent is trained by using the parcel data from day $T$. The parcel volumes for $T+1, T+2, T+3$ are 293208, 322391 and 326332 respectively.

(2) **Proportion**: this is a traditional method used for online parcel assignments. The proportion algorithm relies on the IP solutions from historical parcel data. Here, we collect 30 days' parcel data (total parcel volume > 10 million) before and on day $T$, and solve the offline IP problems. Then, we summarize all the assignments and compute the proportion of parcels assigned to each candidate route. For any incoming parcel, the algorithm randomly assign it to one of the candidate routes with probabilities in proportion to the above proportions computed beforehand.

(3) **PDO**: the primal dual optimization (PDO) [8] is a powerful technique for a wide variety of online problems. In this experiment, we run the PDO algorithm for solving (1), where Lagrangian relaxation is used to control constraint and dual variables would be updated at each iteration. For this algorithm, we use the actual total daily parcel volume as input, which is impossible to acquire in a real scene.

(4) **PPO-PD**: For the reinforcement learning algorithm, Lagrangian relaxation is an effective technique to process soft constraints. Here, we set the reward to negative cost in the Online Assignment MDP framework and use the primal-dual update to control the violation of constraints, which is similar to Chow et al. [10] and leads to the unconstrained problem,

$$\min_{\lambda \geq 0} \max_{\theta} L^{\text{CLIP}}(\theta; \pi) - \sum_{k \in \mathcal{K}} \lambda_k (J_k(\pi) - U_k),$$

where $J_k(\pi)$ represents the capacity from policy $\pi$ for constraint $k$.

Accordingly, performance metrics are:

- Average cost: the total cost of assigned parcels divided by the number of assigned parcels.
- IP gap: the difference between the average cost of the IP solution and the average cost of the compared algorithm's solution, divided by the average cost of the IP solution.
- Violation rate: the number of parcels that violate constraints divided by the total number of parcels. IP solution has zero constraint violation rate for hub capacity constraints and route proportion constraints since IP solution is the optimal solution solved in an offline manner.

Table 1 and 2 show the average cost of parcels, IP gap and violation rate achieved by PPO-OPA, PPO-PD, Proportion, PDO and IP using dataset #1 and dataset #2. PPO-OPA achieves about 0.2-0.3% cost reduction and fewer constraint violation rates than the proportion and PPO-PD algorithms. Moreover, PPO-OPA training by one-day data has almost the same performance as PDO algorithm with known parcel volume. It means that our method is more suitable for real scenarios because we do not need to require or predict the daily parcel volume.

## 6 CONCLUSION

We introduce the online parcel assignment problem, which is aimed at assigning each incoming parcel to a candidate route for delivery, in order to minimize the total cost under consideration of given business constraints. Several challenges exist in this problem, including the large number (beyond $10^5$) of daily parcels to assign, the variability of the number and attributes of parcels and the non-Markovian characteristics of parcel arrival dynamics. We propose the Online Assignment MDP and present a DRL approach named PPO-OPA to tackle this problem. In this approach, Proximal Policy Optimization (PPO) is adopted with a dedicated designed MDP for conducting the online assignment. The actor and reward networks adopt the attention mechanism and parameter sharing to accommodate each incoming parcel with varying numbers and identities of candidate routes. By running experiments on the real datasets of, the proposed approach is validated and compared against other commonly used assignment methods in the logistics industry. The results are quite promising: in the majority of the cases, PPO-OPA obtains similar performance to the primal dual method, but with a weaker assumption that the total parcel volume is not given. Finally, it is noteworthy that our approach actual provides a general framework that can be applied to any other online assignment/allocation problems by designing an appropriate Online Assignment MDP.